\documentclass[letterpaper]{article} 
\usepackage[linesnumbered,ruled,vlined]{algorithm2e}
\usepackage{lipsum} 
\SetAlgoNlRelativeSize{0}

\usepackage{aaai24}  
\usepackage{times}  
\usepackage{helvet}  
\usepackage{courier}  
\usepackage[hyphens]{url}  
\usepackage{graphicx} 
\usepackage{fontawesome5}
\usepackage{amsmath}
\usepackage{natbib}  
\usepackage{caption} 
\usepackage{algorithmicx}
\usepackage{algpseudocode}
\usepackage{rotating}
\usepackage{amssymb}
\usepackage{multirow}
\usepackage{booktabs}
\usepackage{colortbl}
\definecolor{Gray}{gray}{0.9}
\definecolor{lightgreen}{rgb}{0.95, 0.95, 0.95}
\usepackage{dcolumn}
\newcolumntype{d}[1]{D{.}{.}{#1}}
\newcolumntype{B}[3]{>{\boldmath\DC@{#1}{#2}{#3}}c<{\DC@end}}
\usepackage{float}
\usepackage{listings}
\DeclareCaptionStyle{ruled}{labelfont=normalfont,labelsep=colon,strut=off} 
\floatstyle{ruled}
\newfloat{listing}{tb}{lst}{}
\floatname{listing}{Listing}
\title{Learning by Erasing: Conditional Entropy Based Transferable Out-of-Distribution Detection}
\author{
    Meng Xing\textsuperscript{\rm 1,\rm 3},
    Zhiyong Feng\textsuperscript{\rm 1},
    Yong Su\textsuperscript{\rm 2},
    Changjae Oh\textsuperscript{\rm 3}
}
\affiliations{
    \textsuperscript{\rm 1}College of Intelligence and Computing, Tianjin University \\
    \textsuperscript{\rm 2}Tianjin Normal University \\
    \textsuperscript{\rm 3}Centre for Intelligent Sensing, Queen Mary University of London\\
    \{xingmeng, zyfeng, suyong\}@tju.edu.cn, c.oh@qmul.ac.uk
    
}

\usepackage{bibentry}

\begin{document}

\maketitle

\begin{abstract}
Detecting Out-of-distribution (OOD) inputs is crucial to deploying machine learning models to the real world safely. However, existing OOD detection methods require an in-distribution (ID) dataset to retrain the models. In this paper, we propose a Deep Generative Models (DGMs) based transferable OOD detection that does not require retraining on the new ID dataset. We first establish and substantiate two hypotheses on DGMs: DGMs are prone to learn low-level features rather than high-level semantic information; the lower bound of DGM's log-likelihoods is tied to the conditional entropy between the model input and target output. Drawing on the aforementioned hypotheses, we present an innovative image-erasing strategy, which is designed to create distinct conditional entropy distributions for each ID dataset. By training a DGM on a complex dataset with the proposed image-erasing strategy, the DGM could capture the discrepancy of conditional entropy distribution for varying ID datasets, without re-training. We validate the proposed method on the five datasets and show that, without retraining, our method achieves comparable performance to the state-of-the-art group-based OOD detection methods. The project codes will be open-sourced on our project website.
\end{abstract}

\section{Introduction}

Deep neural networks (DNNs) have demonstrated their potential in solving various safety-related computer vision tasks~\cite{add2}, such as autonomous driving \cite{autonomous_driving} and healthcare \cite{healthcare}.
However, DNNs tend to yield confident but incorrect predictions for the distribution-mismatched examples \cite{OOD_begin,add1,add3}, and results in serious consequences, e.g., accidents by autonomous vehicles \cite{Uber-Accident} and incorrect diagnosis in healthcare  \cite{incorrect-diagnosis}.  
Therefore, determining whether inputs are out-of-distribution (OOD) is an important task to safely deploy machine learning models to the real world. 

OOD detection can be performed using labeled data by utilizing output characteristics \cite{OOD_classifier3}, training dynamics \cite{features-pace}, adversarial training \cite{OOD_classifier4,semantic-segmentation,large-scale-semantic}, and metric learning \cite{Adversarial-Attacks, 1-dimension-subspace,classifier_2}.
Since it is time-consuming and laborious to obtain labeled data in real scenarios, as an alternative, Deep Generative Models (DGMs) have been used to capture the sample distribution of In-Distribution (ID) datasets~\cite{input_complexity}.
However, most DGMs-based methods focus on elaborating architectures~\cite{background_statistics, input_complexity}, designing loss functions~\cite{Likelihood-Regret} or statistical models \cite{Distance-Guarantee,new_group}, targeting the specific feature representation or data distribution of ID samples \cite{sample_repairing}, i.e., need retraining to adapt to the normal pattern of the new ID datasets.
This motivates the following unexplored question: \textit{How can we make OOD detection transferable across new ID datasets?}

\begin{figure}[t]
\centering
\begin{tabular}{c@{\hspace{3pt}}c }
 \includegraphics[width=3.8cm]{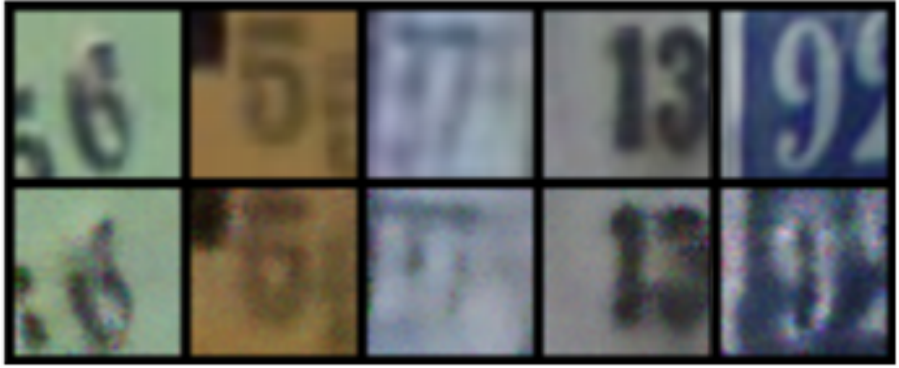}
& \includegraphics[width=3.8cm]{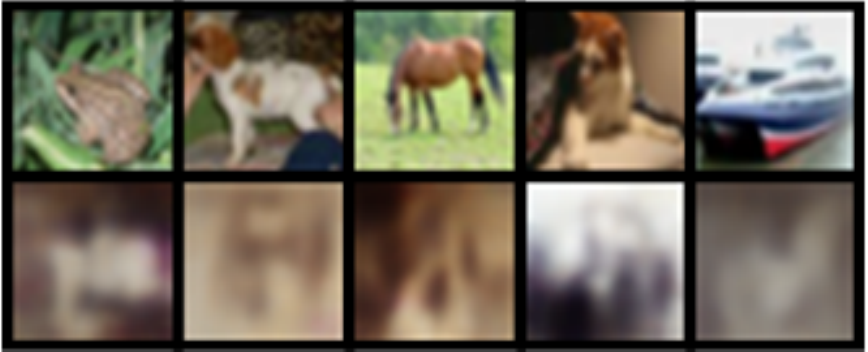}\\
 \footnotesize{CI$\rightarrow$SV (0.85 / 24dB)} 
& \footnotesize{SV$\rightarrow$CI (0.23 / 10dB)}\\
  \includegraphics[width=3.8cm]{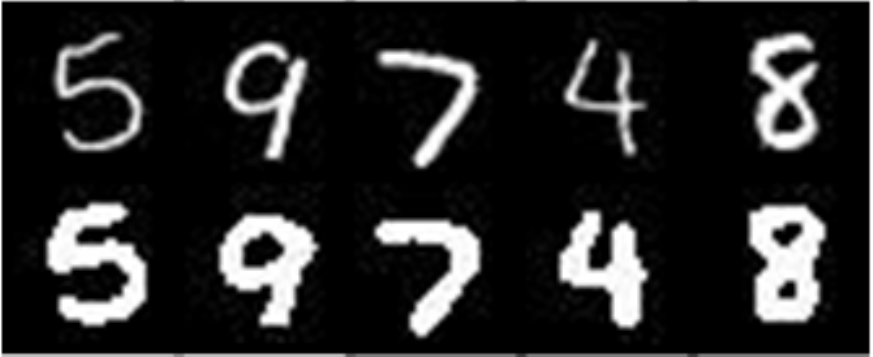}
& \includegraphics[width=3.8cm]{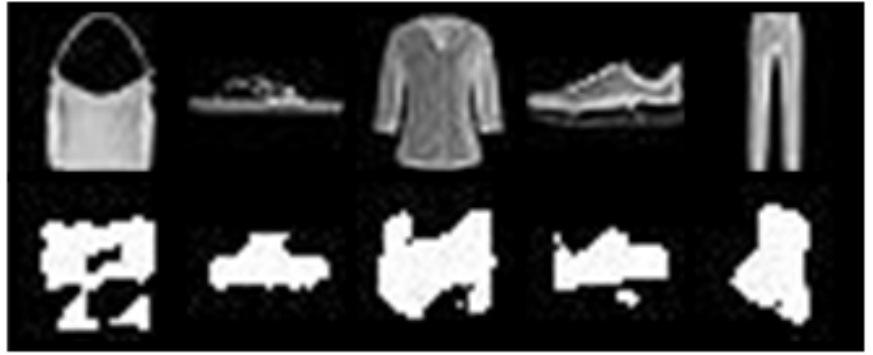}\\
 \footnotesize{Fa$\rightarrow$MN (0.63 / 16dB)}
& \footnotesize{MN$\rightarrow$Fa (0.26 / 9dB)}\\
\end{tabular}
\caption{Original images (top row) and their reconstruction results (bottom row) by a pre-trained DGM. The model pre-trained on CI (CIFAR10) / Fa (FashionMNIST) can reconstruct SV (SVHN) / MN (MNIST) samples well, but not \textit{vice versa}. 
Reconstruction performances on the test set of the target datasets are evaluated with (SSIM$\uparrow$ / PSNR$\uparrow$).}
\label{figure1}
\end{figure}

\begin{figure*}[t]
\centering
\includegraphics[width=16cm]{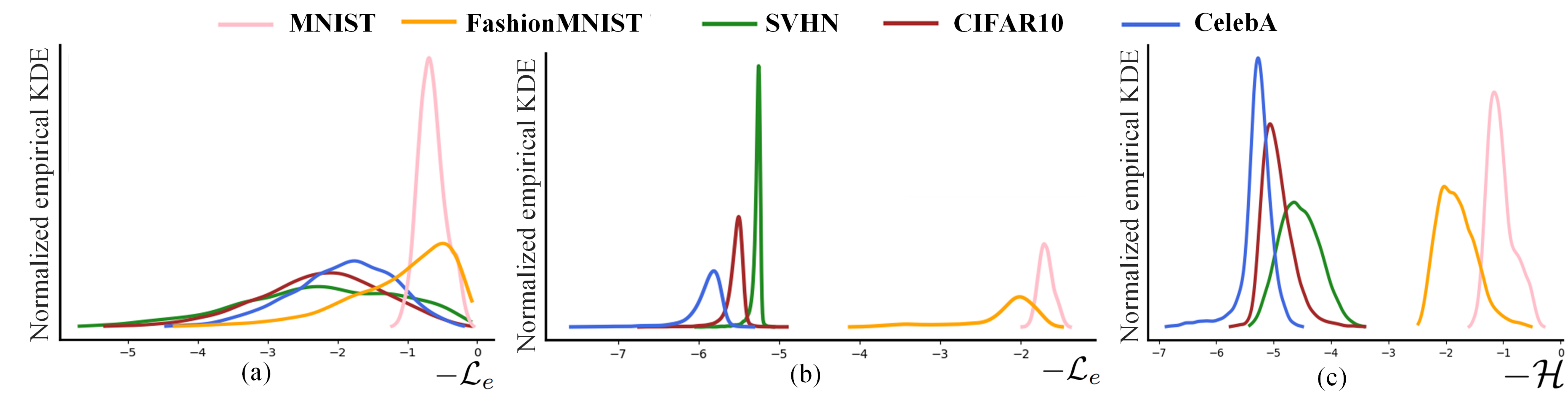}
 \caption{ 
The (a) and (b) are the DGM's negative log-likelihood distribution on different datasets.
The model of (a) is trained by reconstructing the input, while the model of (b) is trained by generating the erased patch based on its surrounding information.
The real negative conditional entropy distribution between the erased patch and its surrounding is given in (c).
The DGM is an auto-encoder proposed in this paper and is trained with ImageNet.
The image size is $32\times32$, and the erased image patch in (b) and (c) is at the center of the image with the size of $16\times16$.
Kernel Density Estimation (KDE) is used to estimate the probability distribution.}
\label{figure2}
\end{figure*}

In this paper, we aim to achieve transferable OOD detection based on the following two key hypotheses:
\textit{1) The DGMs are prone to learn low-level features, rather than semantic information \cite{local_pixel_correlations}.}
Following the experimental setup in \cite{Likelihood-Regret}, we use the Variational Auto-Encoder to reconstruct the input image and show some results comparisons in Figure~\ref{figure1}.
Figure~\ref{figure1} demonstrates that a DGM pre-trained on a complex dataset, which includes diverse semantic categories and a complex image texture, can capture the distribution of simple datasets, but not \textit{vice versa}. 
This means that the DGMs pre-trained on a complex dataset can approach the lower bounds of negative-log-likelihoods of simple datasets without retraining.

\textit{2) In the DGMs, the lower bound of negative-log-likelihoods is determined by the conditional entropy between the model input and target output.}
We give supporting experimental results in Figure~\ref{figure2} and theoretically demonstrate this hypothesis in \textit{Motivation}. 
The log-likelihood distributions of all datasets in Figure~\ref{figure2}(a) are approaching 0 since the conditional entropy between the model input and output is 0 in this experiment setting.
This result explains why traditional DGMs cannot be used directly for OOD detection\cite{input_complexity}.
In contrast, the log-likelihood distributions of five datasets in Figure~\ref{figure2}(b) are significantly different from each other and the distribution discrepancy is consistent with real conditional entropy distribution in Figure~\ref{figure2}(c).
Therefore, we can assign an exclusive conditional entropy distribution for each dataset by designing an appropriate image-erasing strategy, which is an indispensable prerequisite for achieving transferable OOD detection.

Motivated by the proven hypotheses, we propose a novel Conditional Entropy based Transferable OOD detection (CETOOD).
Specifically, we first propose an image-erasing strategy that creates exclusive conditional entropy distribution for different datasets by considering the erased patch and its surrounding information as the content and condition. 
Subsequently, we design the Uncertainty Estimation Network (UEN), which estimates the Maximum A Posteriori of generating the erased patch by reconstructing the surrounding information and generating the erased patch.
Finally, we train the UEN on ImageNet \cite{ImageNet} dataset, affording our model approaching the lower bounds of negative-log-likelihoods on different ID datasets, which reflects their distribution discrepancy of conditional entropy.
In the experiment, we demonstrate that our method achieves comparable performance with state-of-the-art methods in group-based OOD detection.
More importantly, our pipeline drastically curtails the time and memory cost of model deployment due to its transferability and concise network architecture.
In summary, our contributions are as follows:

\begin{itemize}
\item We introduce the concept of conditional entropy into OOD detection for model transferability, and theoretically demonstrate the lower bound of negative-log-likelihoods in DGMs is determined by the conditional entropy between the model input and target output.

\item We propose a transferable OOD detection method (CETOOD), which captures the distribution discrepancy of conditional entropy of different ID datasets to achieve transferable OOD detection.

\item We demonstrate the effectiveness and lightweight of the proposed method through extensive comparisons with state-of-the-art techniques, across different datasets.
\end{itemize}

\section{Related Work}
\label{section2}

Some \textbf{Classifier-based approaches} detect OOD samples by utilizing the statistical characteristic of class probabilities.
Hendrycks \textit{et al.} \cite{OOD_classifier1} propose maximum softmax probability as a baseline for OOD detection in deep neural network (DNN) algorithms, and ODIN~\cite{ODIN} further enhance the performance by using temperature scaling and adding small perturbations on ID inputs. 
Since the distribution of OOD data is not available, some methods have explored using synthesized data from generative adversarial networks (GANs)~\cite{classification-framework-based-OOD-detection3-GAN-classifier-EM-train} or using unlabeled data \cite{OOD_classifier2,Self-Supervised} as auxiliary OOD training data, which allows the model to be explicitly regularized by fine-tuning, producing lower confidence on anomalous examples.
In addition to these softmax-classification-based frameworks, recently, researchers focus on the feature embedding of the model.
With the observation that the unit activation patterns of a particular layer show a significant difference between ID and OOD data, Djurisic \textit{et al.} \cite{classifier_3} utilize feature transformation to generate the OOD score.
Similarly, some methods exploit hyperspherical embeddings \cite{classifier_2} or cosine similarity \cite{classifier_1} between features to promote strong ID-OOD separability.
Despite the promising results, classification-based approaches show limitations on the non-labeled tasks.

As an alternative, most \textbf{DGM-based OOD detection} methods separate the ID and OOD samples by exploiting the inductive bias of DGMs, including background statistics \cite{background_statistics,frequency_background}, inputs complexity \cite{input_complexity} and low-level features \cite{sample_repairing}.
Xiao \textit{et al.} \cite{Likelihood-Regret} propose the Likelihood Regret, which is a log-ratio between the likelihood of input obtained by posteriori distribution and approximated by VAE, to detect OOD samples.
Serr{\`{a}} \textit{et al.} \cite{input_complexity} design a complexity estimate score and utilize the subtraction between negative log-likelihoods and the complexity estimate score to detect OOD inputs.
Kirichenko \textit{et al.} \cite{local_pixel_correlations} prove through experiments that what DGMs learn from images is local pixel correlation and local geometric structure rather than semantic information.
Therefore, Sun \textit{et al.} \cite{sample_repairing} utilizes sample repairing to encourage the generative model to focus on semantics instead of low-level features.
A recent work \cite{DGM-error} has shown that for the point-based OOD detection method, a perfect model can perform worse than a falsely estimated one when the ID and OOD data are overlapped.

Therefore, \textbf{Group-based OOD detection methods} utilize the distribution characteristics of grouped inputs for OOD detection.
Most group-based methods consider either the raw input or a certain representation of samples for OOD detection. 
Nalisnick \textit{et al.} \cite{Typicality} propose an explicit test for typicality employing a Monte Carlo estimate of the empirical entropy.
As an alternative, exploit data representations in the latent space can also be utilized to achieve OOD.
Zhang \textit{et al.} \cite{Distance-Guarantee} find that the representations of inputs in DGMs can be approximated by fitted Gaussian and the distance between the distribution of representations of inputs and prior of the ID dataset can be utilized to detect OOD samples.
Jiang \textit{et al.} \cite{new_group} propose to compare the training and test samples in the latent space of a flow model.
However, these methods require retraining when encountering new ID datasets, which is computationally expensive and time-consuming.

\vspace{-5pt}
\section{Motivation}\label{section3}

In this section, we demonstrate the relationship between the lower bound of DGM's negative-log-likelihoods and the conditional entropy between model input and target output.
We take the grayscale image as an example, which can be extended to RGB easily.

Given an image pair $(A, B)$, we can calculate the uncertainty of random variable $B$ given random variable $A$, i.e., the conditional entropy $H(B \vert A)$ as follows:
\begin{align}\label{equation1}
H(B \vert A) = &- \Sigma_{i = 0}^{N_B} \Sigma_{j = 0}^{N_A} P(A_j, B_i)log(P(B_i \vert A_j)) \notag \\
 = &- \Sigma_{i = 0}^{N_B} P(B_i)log(P(B_i \vert A)) \notag
\end{align}
where $A_j$ and $B_i$ are the pixel value at locations $j$ and $i$ of images $A$ and $B$.
$N_A$ and $N_B$ are the number of pixel of images $A$ and $B$.

For image generation, given the model input $A$, target output $B$ and a pre-trained DGM (parameters: $Z$), the Maximum A Posteriori (MAP) of generating output can be estimated as follows:
\begin{equation}\label{equation2}
MAP = \mathop{\arg\min}\limits_{Z} KL(P(B \vert Z)P(Z) \;\vert\vert\; P(B \vert A)P(A))  \notag
\end{equation}
According to the information bottleneck theory \cite{bottleneck}, the lower bound of the negative-log-likelihoods of DGM can be formulated as follows:
\begin{align}\label{back_e3}
\mathcal{L}_{lower \; bound} & = -(log(P(B \vert A)) + log(P(A))) \notag \\
&= - (\Sigma_{i = 0}^{N_B} log(P(B_i \vert A))/N_B)  - log(P(A)) \notag \\
&= \underbrace{ - (\Sigma_{i = 0}^{N_B} P(B_i)log(P(B_i \vert A)))}_\text{$H(B \vert A)$}  - log(P(A))  \notag
\end{align}
where $P(B_i \vert A)$ is modeled by the pre-trained DGM.

Therefore, given A as input, the conditional entropy between A and B would determine the lower bound of DGM's negative-log-likelihoods.

\begin{figure}[t]
\centering
\includegraphics[width=8.6cm]{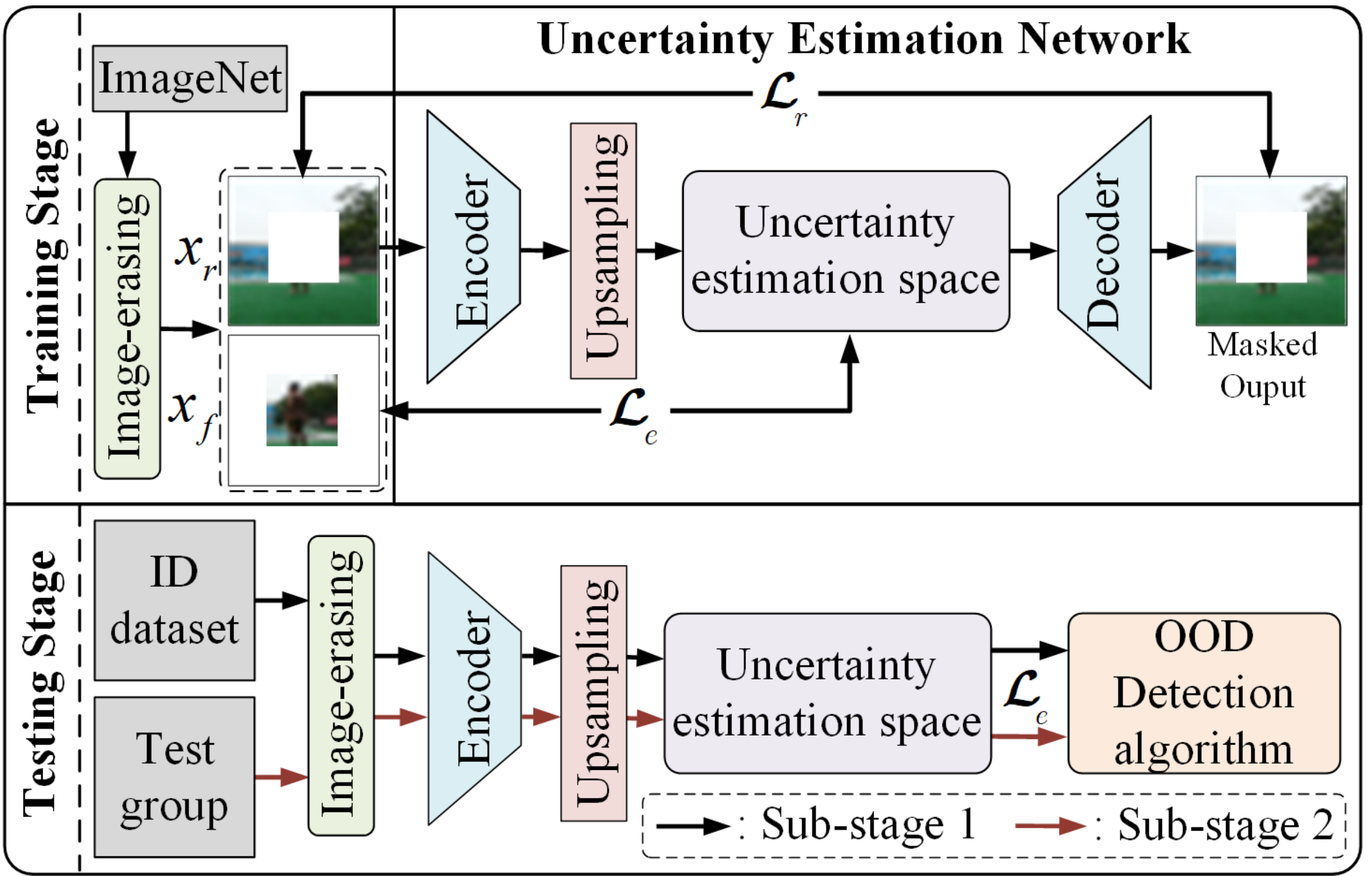}
\caption{The pipeline of the proposed CETOOD. }
\label{figure4}
\end{figure}

\vspace{-2pt}
\section{Method}\label{section4}
The proposed framework consists of the image-erasing strategy, UEN, and OOD detection algorithm, as shown in Figure \ref{figure4}.

\vspace{-2pt}
\subsection{Image-Erasing Strategy}
\label{section4.1}

\begin{figure}[t]
\centering
\includegraphics[width=8cm]{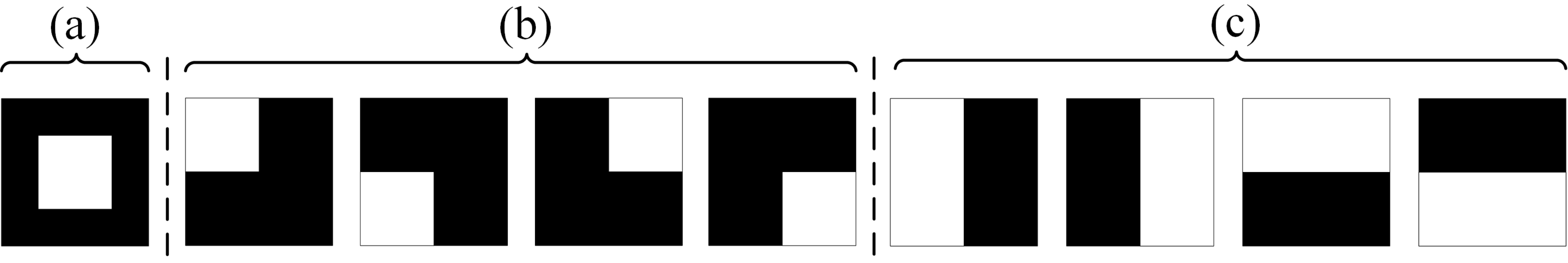}
\caption{The center (a), corner (b) and side (c) of the image is erased, which is indicated with white color.}
\label{figure3}
\end{figure}

To create exclusive conditional entropy distribution for different datasets, we design an image-erasing strategy that divides the image into the erased patch and its surrounding information.
Conditional entropy is a measure of the information difference between the image's erased patch and its surrounding.
Due to semantic differences existing between different datasets, exclusive conditional entropy distributions can be generated for different datasets by erasing the most semantically meaningful regions.
We empirically choose to erase the center of the input, but also propose other erasing strategies for comparison.
The details of the image-erasing strategies are shown in Figure \ref{figure3}.

With the original image $x$, the model input (surrounding information) $x_r$ and output (erased patch) $x_f$ are generated as follows:
\begin{equation}\label{meth_e1}
x_{r} = Mask(x),~~\text{with}~~x_{f}= x - x_{r},
\end{equation}
\noindent
where $Mask(x)$ indicates putting a mask on $x$.

\vspace{-2pt}
\subsection{Uncertainty Estimation Network}
\label{section4.2}
To capture the distribution discrepancy of conditional entropy for different datasets, we propose UEN, a concise auto-encoder, as shown in Figure~\ref{figure4}.
To estimate the MAP of generating the erased patch, UEN needs to calculate the probability of generating the target output directly based on the model input.
We assume that the pixel values at each position of the image conform to a continuous distribution, and all parameters of the distribution depend on the model input.
Inspired by PixelCNN++~\cite{PixelCNN++}, we use the mixed logistic distribution as the above continuous distribution and name the feature space on which the parameters of the distribution depend as uncertainty estimation space, $Z$:
\begin{equation}\label{meth_e2}
  Z \sim \sum_{k=1}^{K} \pi_k \frac{e^{-(x-\mu_k)/\gamma_k}}{\gamma_k (1+e^{-(x-\mu_k)/\gamma_k})^2},
\end{equation}
where $K$ is the number of components in the mixed logistic distribution, $\pi_k$ is the weight of each component, $\mu_k$ and $\gamma_k$ are the shape and position parameters of the logistic distribution, respectively ($\pi_k$, $\mu_k$ and $\gamma_k$ are learnable parameters, where $k = \{1 , 2, \ldots, K\}$).

Given the erased patch, $x_f$, the likelihood of each discretized pixel value can be directly calculated as follows:
\begin{equation}\label{meth_e3}
 P(x_f \vert Z)=\sum_{k=1}^{K} \pi_k [    \sigma (\frac{x_f+ \frac{1}{255}-\mu_k} {\gamma_k })- \sigma (\frac{x_f-\frac{1}{255}-\mu_k} {\gamma_k })],
\end{equation}
where $\sigma(\cdot)$ is the sigmoid function, and we set $K$ to $10$ in this paper.
For RGB images, we only allow linear dependence between three-channel pixel values.

The encoder consists of three parallel multi-layer convolutional branches with different kernel sizes. Upsampling layers are designed to ensure the size of the feature map in the uncertainty estimation space is consistent with the input. The deep feature in uncertainty estimation space is further mapped into the image domain by a decoder.

To ensure that no surrounding information is lost in the process of constructing the uncertainty estimation space, i.e, $P(x_f \vert Z) \approx P(x_f \vert x_r)$, we design the reconstruction loss, $\mathcal{L}_{r}$, as follows:
\begin{equation}\label{meth_e4}
 \mathcal{L}_{r} = \|  x_r - o_r  \|^2,
\end{equation}
where $o_r = Mask(o)$ is the masked output and $o$ is the model output.

To highlight the distribution discrepancy, the generation loss, $\mathcal{L}_{e}$, which measures the posterior probability of generating the erased patch, is presented as:
\begin{align}\label{meth_e5}
\mathcal{L}_{e}   & = -log_2[P(x_f \vert Z)] \notag \\
& = - \sum_{i=1}^{N}log_2[P(x_{fi} \vert Z)]/N_f,
\end{align}
where $i$ is the pixel location, $x_{fi}$ is the pixel value at location $i$ and $N_f$ is the number of pixels in the erased patch, $x_{f}$.

$\mathcal{L}_{e}$ encourages UEN to narrow the log-likelihood distribution gap between the samples that contain similar semantic discrepancies.
The final loss function is as follows:
\begin{equation}\label{meth_e6}
 \mathcal{L}_{total} = \lambda \mathcal{L}_{r} + (1-\lambda)\mathcal{L}_{e},
\end{equation}
where $\lambda$ is used to balance the effect between $\mathcal{L}_{r}$ and $\mathcal{L}_{e}$.

\begin{algorithm}[t]
\caption{OOD Detection Algorithm}
\label{alg1}
 \begin{algorithmic}[1]
\Require $Z$: pre-constructed uncertainty estimation space;
 $X^* = \{x^*_1 , x^*_2, \ldots x^*_N\}$: all of ID samples;
 $X = \{x_1, x_2, \ldots x_m\}$: a batch of of test samples;
 $Mask()$: the function of erasing image patch;
 $t$: threshold.
\State $i \gets 1$
\State \textbf{while} {$i  \leq N$}
\State \hspace*{0.2in}$x^*_{if} = x^*_i - Mask(x^*_i)$
\State \hspace*{0.2in}$L^*[i] = \mathcal{L}_{e}(x^*_{if} \vert Z); i \gets i + 1$
\State \textbf{end while}
\State $P(L^*)=KDE(L^*)$
\State \textbf{while} {$testset \neq \varnothing$}
\State \hspace*{0.2in}$j \gets 1$
\State \hspace*{0.2in}\textbf{while} {$j \leq m$}
\State \hspace*{0.4in}$x_{jf} = x_j - Mask(x_j)$
\State \hspace*{0.4in}$L[j] = \mathcal{L}_{e}(x_{jf} \vert Z);j \gets j + 1$
\State \hspace*{0.2in}\textbf{end while}
\State \hspace*{0.2in}$P(L)=KDE(L)$
\State \hspace*{0.2in}$k=KL(P(L)\|P(L^*))$
\State \hspace*{0.2in}\textbf{if} {$k>t$}
\State \hspace*{0.4in}return $X$ is out-of-distribution data.
\State \hspace*{0.2in}\textbf{else}
\State \hspace*{0.4in}return $X$ is in-distribution data.
\State \hspace*{0.2in}\textbf{end if}
\State \hspace*{0.2in}reload $X$
\State \textbf{end while}
\end{algorithmic}
\end{algorithm}

\begin{table*}[t]
\small
\begin{center}
\begin{tabular}{cll|rrrrrrrr}
\bottomrule
\rowcolor{Gray}
 \multicolumn{3}{c}{Methods} \vline &  \multicolumn{2}{c}{DOCR-TC-M} & \multicolumn{2}{c}{Ty-test} & \multicolumn{2}{c}{RF-GM}& \multicolumn{2}{c}{Ours}\\
\rowcolor{Gray}
\multicolumn{3}{c}{{(\textit{\footnotesize Retraining})}} \vline &  \multicolumn{2}{c}{(\textit{\footnotesize Required})} &\multicolumn{2}{c}{(\textit{\footnotesize Required})} & \multicolumn{2}{c}{(\textit{\footnotesize Required})} & \multicolumn{2}{c}{\textbf{\textit{\footnotesize (Not required)}}}\\
\hline
\cline{1-11}
GS&\multicolumn{1}{c}{ID}&\multicolumn{1}{c}{OOD} \vline &  AUROC & AUPR & AUROC & AUPR & AUROC & AUPR & AUROC & AUPR\\
\hline
\multirow{8}{*}{\textbf{5}}  &  \multicolumn{1}{l}{MNIST} & \multicolumn{1}{l}{F-MNIST} \vline & - & - & - & - & -&-&99.2 & 97.2\\
\cline{2-11}
&\multicolumn{1}{l}{F-MNIST} & \multicolumn{1}{l}{MNIST}\vline & \textbf{100.0} & \textbf{100.0} & 95.5 & 92.1 & 99.0&99.0&93.8 & 89.5\\
\cline{2-11}
&\multirow{2}{*}{SVHN} & \multicolumn{1}{l}{CIFAR10}\vline& 99.7 & 99.7 & \textbf{100.0} & \textbf{100.0} &89.0&93.0& 99.2 & 97.5\\
&& \multicolumn{1}{l}{CelebA} \vline& \textbf{100.0} & \textbf{100.0} & 100.0 & 100.0 & 92.0&94.0&98.9 & 96.0\\
\cline{2-11}
&\multirow{2}{*}{CelebA} & \multicolumn{1}{l}{CIFAR10} \vline& 91.6 & 91.9 & 5.7 & 31.2 &\textbf{92.0}&\textbf{93.0}& 91.2 & 86.8\\
&& \multicolumn{1}{l}{SVHN} \vline& \textbf{100.0} & \textbf{100.0} & 83.1 & 80.1 &97.0&96.0& 91.2 & 87.2\\
\cline{2-11}
&\multirow{2}{*}{CIFAR10} & \multicolumn{1}{l}{SVHN} \vline& 99.0 & \textbf{99.6} & 98.6 & 99.3 & 88.0&83.0&\textbf{99.7} & 85.6\\
&& \multicolumn{1}{l}{CelebA} \vline& \textbf{100.0} & \textbf{100.0} & 100.0 & 100.0 & 76.0&77.0&98.6 & 92.6\\
\hline
\multirow{8}{*}{\textbf{10}} & MNIST & \multicolumn{1}{l}{F-MNIST}\vline & - & - & - & - &-&-& 100.0 & 100.0\\
\cline{2-11}
&F-MNIST & \multicolumn{1}{l}{MNIST}\vline & \textbf{100.0} & \textbf{100.0} & 99.4 & 99.3 & 99.0&99.0&99.1 & 96.3\\
\cline{2-11}
&\multirow{2}{*}{SVHN} & \multicolumn{1}{l}{CIFAR10} \vline& 100.0 & 100.0 & 100.0 & 100.0 & 95.0& 98.9&\textbf{100.0} & \textbf{100.0}\\
&& \multicolumn{1}{l}{CelebA} \vline& 100.0 & 100.0 & 100.0 & 100.0 & 98.0&99.0& \textbf{100.0} & \textbf{100.0}\\
\cline{2-11}
&\multirow{2}{*}{CelebA} & \multicolumn{1}{l}{CIFAR10} \vline& 99.2 & \textbf{99.3} & 0.9 & 30.7 & 98.0& 99.0&\textbf{99.4} & 93.9\\
&& \multicolumn{1}{l}{SVHN} \vline& \textbf{100.0} & \textbf{100.0} & 91.6 & 90.5 & 100.0& 100.0&99.0 & 94.6\\
\cline{2-11}
&\multirow{2}{*}{CIFAR10} & \multicolumn{1}{l}{SVHN} \vline & \textbf{100.0} & \textbf{100.0} & 99.9 & 100.0 &99.0&98.0& 99.3 & 99.7\\
&& \multicolumn{1}{l}{CelebA}\vline & \textbf{100.0} & \textbf{100.0} & 100.0 & 100.0 & 89.0&90.0&99.9 & 99.9\\
\toprule
\end{tabular}

\caption{The OOD detection results with different group size (GS) on five different datasets. Unlike other methods, our transferable method does not require retraining on the ID dataset.}
\vspace{-10pt}
\label{Experiments_t1}
\end{center}
\end{table*}

\vspace{-3pt}
\subsection{OOD Detection Algorithm}\label{section4.3}

Algorithm~\ref{alg1} shows the proposed OOD detection using the pre-trained uncertainty estimation network. Given all ID samples $X^* = \{x^*_1, x^*_2,\ldots x^*_N\}$ and image-erasing strategy $Mask()$, we first utilize Kernel Density Estimation (KDE) to obtain the distribution of log-likelihood for ID dataset.
Then, in the same way, given a set of test samples $X = \{x_1, x_2, \ldots, x_n\}$, we estimate the distribution of log-likelihood on the test group. 
Finally, we measure the estimated total correlation between the test group and the ID samples by using KL-divergence, and determine the test group as the OOD data if there exists a significant distribution discrepancy.

\vspace{-6pt}
\section{Experiments}
\label{section5}

\subsection{Implementation Details}\label{section5.1}
All three parallel encoder branches consist of multiple convolution and upsampling layers with different kernel sizes (3$\times$3, 5$\times$5 and 7$\times$7).
A shared convolutional layer with a kernel size of 1$\times$1 is utilized to transform the features from $3$ parallel encoder branches into uncertainty estimation space.
The decoder consists of two convolutional layers with kernel size of 3$\times$3.
We set the batch size and learning rate to 64 and $10^{-5}$, respectively.
$\lambda$ is empirically set to 0.8.
We trained the network for 250 epochs, taking about 48.29 hours.
We conduct all experiments on a single NVIDIA GPU 3080 that follows the experimental setup of the baseline methods.

\vspace{-3pt}
\subsection{Experimental Setting}\label{sec4.2}
\subsubsection{Datasets}\label{sec4.2.1}
We train our model on ImageNet32 \cite{ImageNet} and validate our model on different ID datasets, including MNIST \cite{MNIST}, FashionMNIST \cite{Fashion-MNIST}, SVHN \cite{SVHN}, CelebA \cite{celeA} and CIFAR10 \cite{CIFAR10}.
All the inputs are resized to 32$\times$32 to fit the input size of UEN.
We transform the grayscale image into an RGB image by replicating the channel.
\vspace{-2pt}
\subsubsection{Metrics}\label{sec4.2.2}
We use threshold-independent metrics: the area under the receiver operating characteristic curve (AUROC)~\cite{AUROC} and the area under the precision-recall curve~(AUPR) to evaluate our method.
We consider OOD data and ID data as positive and negative ones for detection, respectively.
Unless noted otherwise, we calculate the False Positive Rate (FPR) of the detector when the threshold is set at 95\% TPR.
We randomly select 10k samples from the test set of the target dataset.
We generate test sample groups according to group size $gs$.
For the fair comparison, we generate the test set $2$ times and test groups $5$ times then report the averaged result.

\vspace{-3pt}
\subsection{OOD Detection}
To evaluate the robustness of our method, we utilize five different datasets as ID datasets and test each of them on one (MNIST or FashionMNIST) or three (SVHN, CelebA and CIFAR10) different disjoint OOD datasets.
The obtained performance for OOD detection and comparison with three baselines including the Ty-test \cite{Typicality}, DOCR-TC-M \cite{Distance-Guarantee} and RF-GM \cite{new_group} are shown in Table~\ref{Experiments_t1}.
We utilize the three methods as our baselines as they outperform other existing group-based OOD detection methods.
As shown in Table~\ref{Experiments_t1}, our method can achieve competitive performance compared with the SOTA methods.
Our method achieves higher AUROC compared to  RF-GM across various detection scenarios, especially, our method outperforms RF-GM 22.6\% AUROC when detecting CelebA from CIFAR10 with $5$ as group size.
Likewise, our method shows 0.7\% higher AUROC compared to DOCR-TCM when detecting SVHN from CIFAR10 with $5$ as group size.
Notably, compared to the baseline methods, our framework does not require retraining when deployed on new ID datasets.

\begin{table*}[t]
\small
\centering
\setlength\tabcolsep{2mm}{
\begin{tabular}{m{1.3cm}m{1.3cm}|>{\centering}m{1cm}%
>{\centering\arraybackslash}m{1cm}%
>{\centering\arraybackslash}m{1.1cm}%
>{\centering\arraybackslash}m{1.1cm}%
>{\centering\arraybackslash}m{1.1cm}%
>{\centering\arraybackslash}m{1.1cm}%
>{\centering\arraybackslash}m{1.1cm}%
>{\centering\arraybackslash}m{1.1cm}%
>{\centering\arraybackslash}m{1.1cm}%
}
\bottomrule

\multirow{2}{*}{ID} & \multirow{2}{*}{OOD}  & \multicolumn{3}{c}{(a) Group size} \vline & \multicolumn{3}{c}{(b) Erasing strategy} \vline &\multicolumn{3}{c}{(c) Erasing strategy ($\mathcal{H}$)}\\
\cline{3-11}
&  & \multicolumn{1}{c}{20} &\multicolumn{1}{c}{50} & \multicolumn{1}{c}{100} \vline   & \multicolumn{1}{c}{corner} & \multicolumn{1}{c}{side} & \multicolumn{1}{c}{center} \vline& \multicolumn{1}{c}{corner($\mathcal{H}$)} & \multicolumn{1}{c}{side($\mathcal{H}$)} & \multicolumn{1}{c}{center($\mathcal{H}$)}\\
\hline
MNIST & F-MNIST & \multicolumn{1}{c}{100.0} & \multicolumn{1}{c}{100.0} & \multicolumn{1}{c}{100.0} \vline  &  \multicolumn{1}{c}{95.2} & \multicolumn{1}{c}{99.7} & \multicolumn{1}{c}{100.0} \vline &\multicolumn{1}{c}{99.7} & \multicolumn{1}{c}{99.9} & \multicolumn{1}{c}{99.9}\\
\hline
F-MNIST & MNIST& \multicolumn{1}{c}{99.9} &\multicolumn{1}{c}{100.0}& \multicolumn{1}{c}{100.0} \vline  & 91.4 & 97.2 & \multicolumn{1}{c}{99.1} \vline & 94.2 & 97.7 & 99.6\\
\hline
\multirow{2}{*}{SVHN} & CelebA& 100.0 &100.0& \multicolumn{1}{c}{100.0} \vline  & 100.0 &  100.0 & \multicolumn{1}{c}{100.0} \vline &99.7 & 99.7 & 99.9\\
& CIFAR10 & 100.0 &100.0& \multicolumn{1}{c}{100.0} \vline  & 99.9  & 100.0 & \multicolumn{1}{c}{100.0} \vline & 99.5 & 99.1 & 99.9 \\
\hline
\multirow{2}{*}{CelebA } & SVHN & 99.9 & 100.0& \multicolumn{1}{c}{100.0} \vline & 89.4 & 90.1 & \multicolumn{1}{c}{99.0} \vline & 100.0 & 100.0 & 100.0\\
& CIFAR10 & 99.8 &100.0 & \multicolumn{1}{c}{100.0} \vline & 64.8 & 63.7 &  \multicolumn{1}{c}{99.4} \vline & 56.3 & 54.3 & 86.3\\
\hline
\multirow{2}{*}{CIFAR10} & SVHN & 99.6 &99.9 & \multicolumn{1}{c}{100.0} \vline & 90.3 & 95.2 & \multicolumn{1}{c}{99.3} \vline & 100.0 & 100.0 & 100.0\\
& CelebA & 100.0  &100.0& \multicolumn{1}{c}{100.0} \vline & 59.3 & 58.7 & \multicolumn{1}{c}{99.9} \vline & 51.2 & 51.0 & 89.6\\
\hline
\multirow{2}{*}{ID} & \multirow{2}{*}{OOD}  & \multicolumn{3}{c}{(d) Loss function} \vline& \multicolumn{6}{c}{(e) Training set}\\
\cline{3-11}
 &  &\multicolumn{1}{c}{$\mathcal{L}_{e}$} &  \multicolumn{1}{c}{$\mathcal{L}_{r}$} &\multicolumn{1}{c}{$\mathcal{L}_{total}$}\vline & \multicolumn{1}{c}{MNIST} & \multicolumn{1}{c}{F-MNIST} & \multicolumn{1}{c}{SVHN} & \multicolumn{1}{c}{CelebA} & \multicolumn{1}{c}{CIFAR10}  & \multicolumn{1}{c}{ImageNet}\\
\hline
MNIST & F-MNIST & 99.7 & 99.9 &\multicolumn{1}{c}{100.0}\vline & 100.0 & 99.2  & 99.9 & 100.0 &99.8&100.0  \\
\hline
F-MNIST & MNIST & 93.3 & 98.4 &\multicolumn{1}{c}{99.1} \vline & 99.7 & 97.9  & 97.7 & 98.5 & 98.1 & 99.1 \\
\hline
\multirow{2}{*}{SVHN} & CelebA& 61.4 & 77.6 &\multicolumn{1}{c}{100.0}\vline & 61.6 & 74.6  & 100.0 & 49.1 & 100.0 & 100.0 \\
& CIFAR10 & $62.0$ & 76.1 &\multicolumn{1}{c}{100.0}\vline & 49.2 & 90.2  & 99.9 & 99.9 & 78.0 & 100.0  \\
\hline
\multirow{2}{*}{CelebA} & SVHN& 64.1 & 82.0 &\multicolumn{1}{c}{99.0}\vline& 88.6 & 71.8  & 97.3 & 58.5 & 98.4 &99.0 \\
& CIFAR10 & 70.9 & 79.5 &\multicolumn{1}{c}{99.4}\vline &  74.7 & 65.2 & 77.5 & 90.5 & 97.8 & 99.4 \\
\hline
\multirow{2}{*}{CIFAR10} & SVHN& 62.8 & 78.1 &\multicolumn{1}{c}{99.3}\vline& 75.0 & 88.9  & 96.9 & 99.0 & 68.4  & 99.3\\
& CelebA& 68.1 & 77.5 &\multicolumn{1}{c}{99.9}\vline & 46.4 & 65.1  & 63.2 & 68.4 & 99.6 & 99.9 \\
\toprule
\end{tabular}
\caption{ Model performance with different hyperparameters and training variations (group size for b-e is 10).}
\label{Experiments_t2}
\vspace{-1em}}
\end{table*}

\vspace{-4pt}
\subsection{Deployment Cost Analysis}\label{S-T}
\begin{figure}[t]
\vspace{-0.5em}
\centering
\includegraphics[width=8cm]{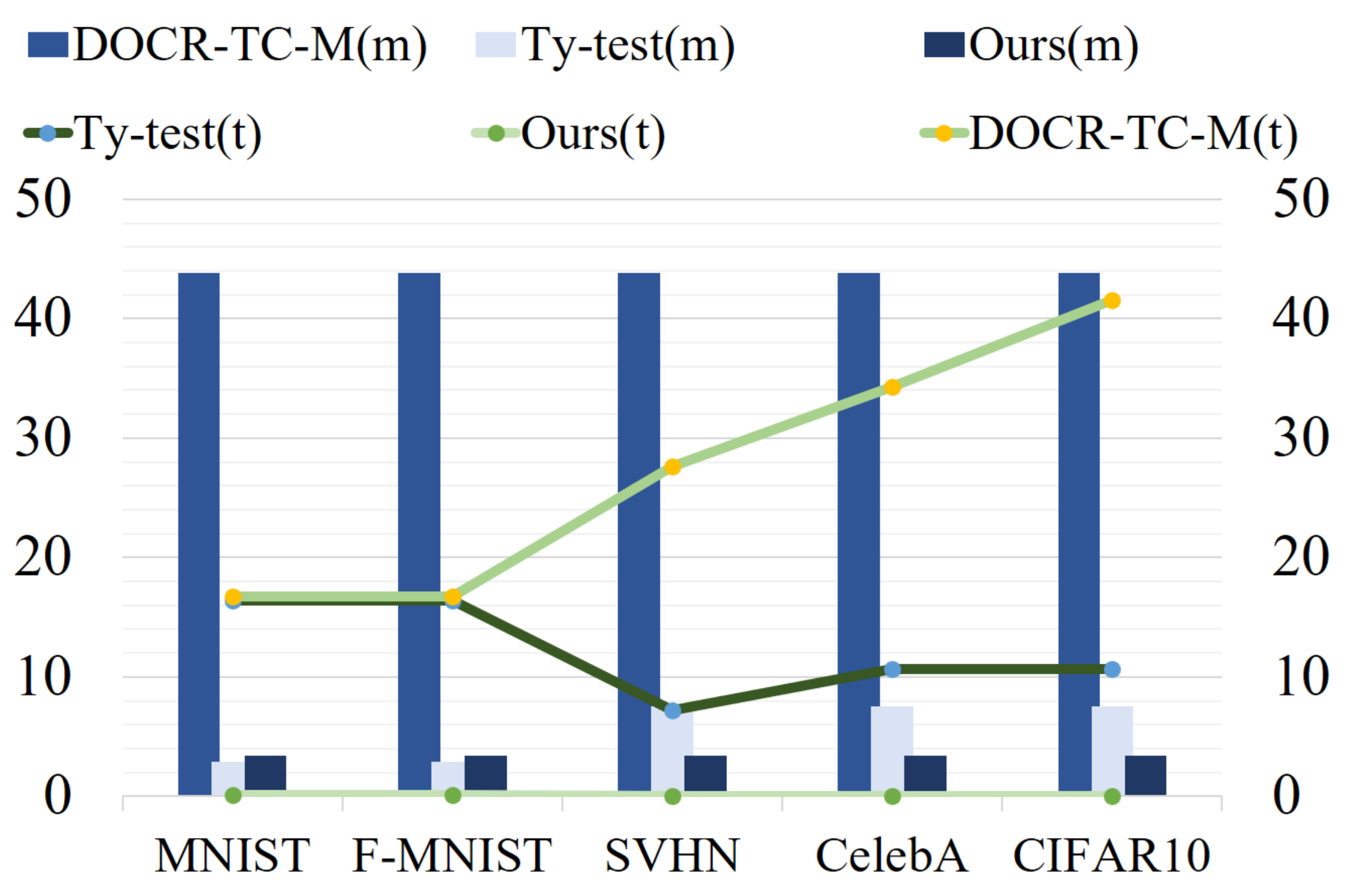}
\caption{The time ($t$, hours) and space ($m$, FLOPs) complexity comparison between our method and the baseline approaches.}
\label{Experiments_f1}
\vspace{-1em}
\end{figure}

In order to comprehensively analyze the performance of our model, we compare training time and memory cost of network parameters of our approach with that of the baseline methods.
Due to both DOCR-TC-M and RF-GM are based on the flow model, we choose the DOCR-TC-M with better performance as a baseline.
The experiment settings for DOCR-TC-M \cite{Distance-Guarantee} and Ty-test \cite{Typicality} are consistent with the original papers.
The training time and memory cost comparison are shown in Figure~\ref{Experiments_f1}.
Our method does not require retraining and only needs to calculate the DGM's likelihood distribution of the new ID dataset in the testing stage.
Therefore, the time cost of model deployment can be greatly reduced.
Our model needs 48.3 hours to be pre-trained on ImageNet, which is still less than the time cost of training the baseline methods on all ID datasets.
In addition, the space complexity comparison in Figure~\ref{Experiments_f1} shows that the memory cost of our model is significantly lower than the baseline methods.

\vspace{-2pt}
\subsection{Ablation Study}

\subsubsection{Effect of group size}\label{sec4.5.1}

Table~\ref{Experiments_t2}(a) reports the model performance with different group sizes.
Experiment results demonstrate that the group size only has a slight impact on model performance and it is sufficient to ensure the performance of the model with the group size higher than $5$.

\vspace{-2pt}
\subsubsection{Effect of image-erasing strategy}\label{sec4.5.3}

To analyze the effect of the image-erasing strategy, we use three image-erasing strategies to train the model.
Note that the same image-erasing strategy is applied to both training the model and OOD detection.
For the image-erasing strategy with different variations, we calculate the average of all the variations.
We tabulate the model performance in Table \ref{Experiments_t2}(b), denoted as corner, side and center.
The experimental results indicate that the center strategy can significantly improve the detection performance in some scenarios (CelebA versus CIFAR10 and CIFAR10 versus CelebA), but only slight performance improvement is observed in other scenarios.
To explore the reasons for poor robustness in performance improvement, we feed the real conditional entropy under different image-erasing strategies into Algorithm ~\ref{alg1} and calculate the OOD detection results, as shown in Table \ref{Experiments_t2}(c).
The experimental results show that in the scenarios with slight performance improvement, the corner and side strategies can create exclusive conditional entropy distributions for different datasets.
The above experimental results support hypothesis 2), i.e., the detection performance of the model depends on the conditional entropy distribution discrepancy between different datasets.
See the appendix for more intuitive visualization results.

\begin{figure}[t]
\centering
\includegraphics[width=8cm]{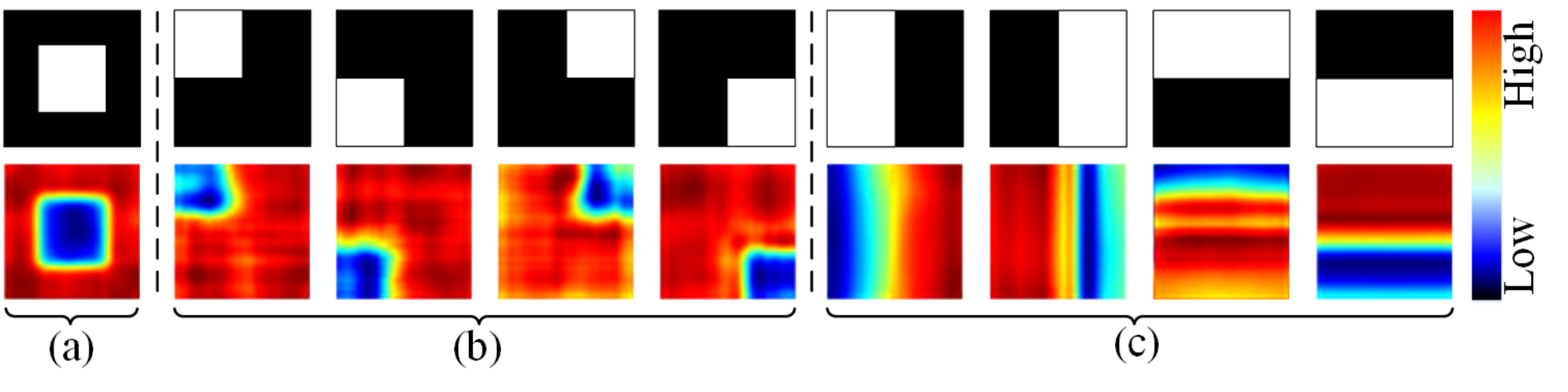}
\caption{Top: Different image-erasing strategies. Bottom: The averaged likelihood heatmaps of all CIFAR10 test samples generated by our model with image-erasing strategies.}
\label{Experiments_f4}
\end{figure}

In addition, as the $\mathcal{L}_{e}$ encourages UEN to narrow the log-likelihood distribution gap between the samples with similar semantic discrepancies, the detection performance of the model should be superior to the detection results based on real conditional entropy.
However, in some experimental settings, the experimental results do not match expectations. 
For example, when detecting SVHN from CIFAR10, the performance of the model decreases compared to the detection results based on real conditional entropy, and the performance degradation caused by different image-erasing strategies varies significantly.
To investigate the impact of the image-erasing strategy on the conditional entropy capturing, we presented the averaged likelihood heatmaps of all samples in the CIFAR10 dataset under different image-erasing strategies, as shown in Figure~\ref{Experiments_f4}.
The expected experimental results should be like Figure~\ref{Experiments_f4}(a), the blue region in the bottom heatmap is aligned with the white region in the top sketch map, which indicates the information discrepancy between the erased patch and its surrounding can be captured by the proposed model effectively.
However, in the corner (Figure~\ref{Experiments_f4}(b)) and side (Figure~\ref{Experiments_f4}(c)) strategies, the blue regions in the heatmaps are much smaller than their corresponding white regions.
The results demonstrate that the model with these two erasing strategies could generate partial information about the target output.
In other words, the model's ability to capture conditional entropy is affected by the value of conditional entropy, the larger the better.

\vspace{-2pt}
\subsubsection{Effect of loss functions}\label{sec4.5.2}

\begin{figure}[h]
\centering
\includegraphics[width=7.5cm]{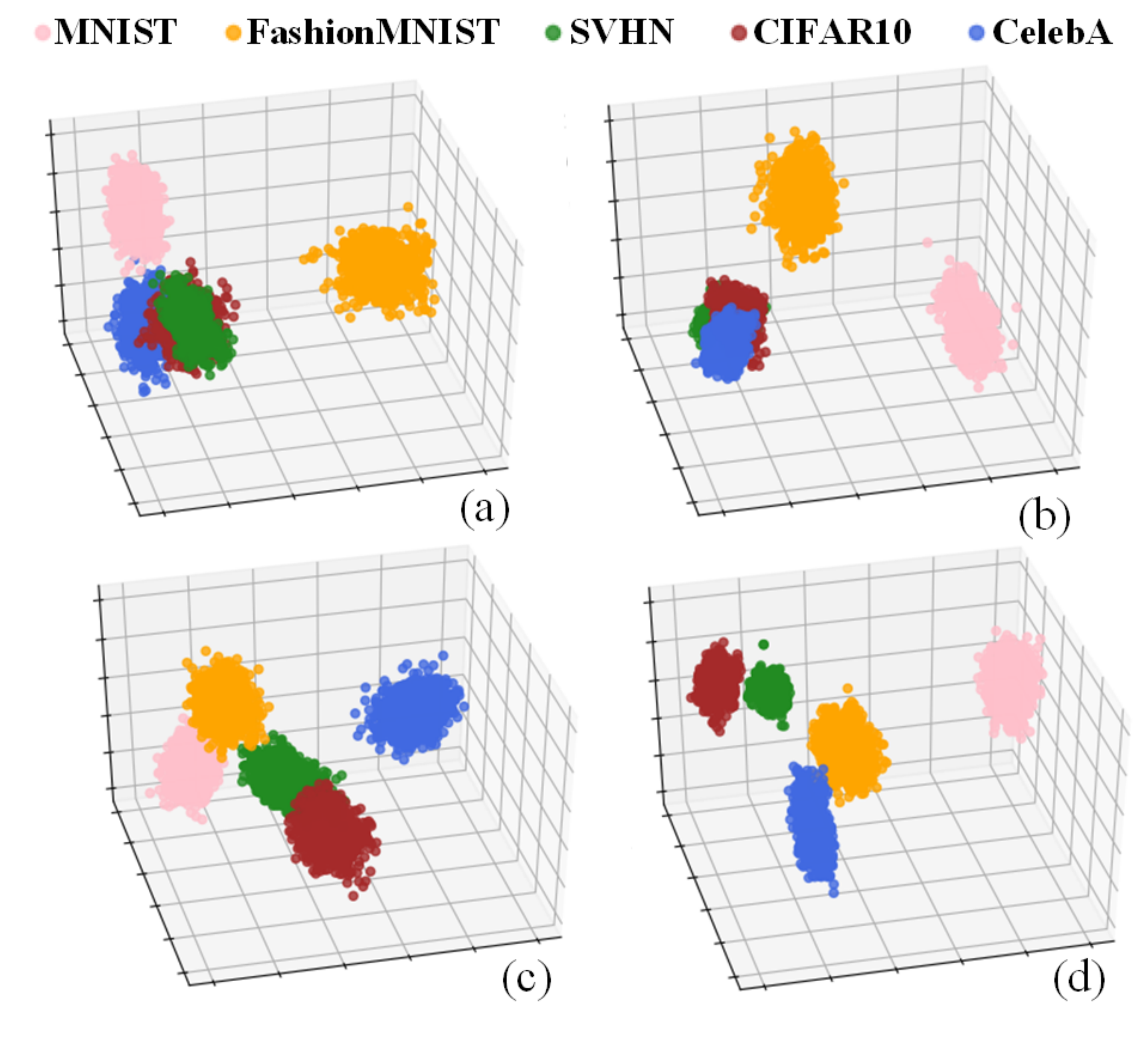}
\caption{Feature visualization of 1000 samples from CIFAR10, extracted from the uncertainty estimation space (Z). The model is trained with $\mathcal{L}_{e}$ (b), $\mathcal{L}_{r}$ (c) and $\mathcal{L}_{total}$ (d), with center image-erasing. The control experiment is training the model with $\mathcal{L}_{total}$, without image-erasing.}
\label{meth_f3}
\end{figure}

We show the quantitative comparison results of different model objectives in Table~\ref{Experiments_t2}(d).
We also show the feature visualization of samples when the model is trained with different model objectives in Figure~\ref{meth_f3}.
Experimental results in Table~\ref{Experiments_t2}(d) show that the performance of $\mathcal{L}_{r}$ consistently outperforms $\mathcal{L}_{e}$ across different datasets, which demonstrates 
ensure $P(x_f \vert Z) \approx P(x_f \vert x_r)$ plays a major role in capturing the inter-dataset distribution discrepancy (the conclusion is consistent with the results in Figure~\ref{meth_f3} (b) and (c)).
In addition, the comparison between Figure~\ref{meth_f3} (c) and (d) shows that $\mathcal{L}_{e}$ reduces the distribution variance of each dataset, thus further increasing the distribution discrepancy.
The results demonstrate that $\mathcal{L}_{r}$ ensures the model captures the condition (surrounding information) of the conditional entropy, while $\mathcal{L}_{e}$ encourages the model to capture the content (erased patches), and the complementarity between them helps to accurately capture the conditional entropy.

\vspace{-2pt}
\subsubsection{Effect of training set}
To analyze the effect of the training set, we train our model using training sets of 6 datasets, including MNIST, F-MNIST, SVHN, CelebA, CIFAR10 and ImageNet. As shown in Table~\ref{Experiments_t2}(e), the results show that the model performance improves with the increases in training data's complexity and achieves optimal performance on ImageNet.
The experimental results support hypothesis 1), i.e., the DGMs learn low-level features rather than semantic information.
As the same image-erasing strategy is used among 6 experiment settings, the experimental results also demonstrate that conditional entropy capturing is affected by the complexity of training data.
Highly complex training data helps the model better capture the conditional entropy of generating the erased patch from its surrounding.
See the appendix for more intuitive visualization results.

\vspace{-2pt}
\subsubsection{Limitation}

\begin{table}[h]
\small
\begin{center}
\begin{tabular}{ll|cccc}
\bottomrule
\multirow{2}{*}{ID} & \multirow{2}{*}{OOD} & \multicolumn{2}{c}{Ours} &\multicolumn{2}{c}{Ours ($\mathcal{H}$)}\\
\cline{3-6}
\multicolumn{2}{c}{}\vline &AUROC &AUPR &AUROC&AUPR\\
\hline
C10&C100 & 64.1&52.3&50.1&49.4\\
C100&C10&62.2&53.4&48.4&49.8\\
\toprule
\end{tabular}
\caption{The OOD detection results on CIFAR10 (C10) and CIFAR100 (C100), the group size is 10.}
\label{Experiments_t3}
\end{center}
\end{table}

To further explore the potential of CETOOD, we utilize the model that is pre-trained on ImageNet to distinguish CIFAR100 \cite{CIFAR10} and CIFAR10.
We also feed the real conditional entropy into Algorithm 1 for OOD detection.
The center image-erasing strategy is used in both experiments.
The reason for poor model performance in Table \ref{Experiments_t3} is that the current image-erasing strategy cannot create exclusive conditional entropy distribution for CIFAR10 and CIFAR100.
The performance improvement compared with the detection results of real conditional entropy proves that our model has the ability to capture conditional entropy. 
For more detailed experimental results, see Appendix.

\section{Conclusion}
We proposed a method to perform transferable OOD detection by leveraging the concept of conditional entropy to OOD detection.
We first validated two hypotheses: The DGMs are prone to learn low-level features rather than semantic information. In the DGMs, the lower bound of negative-log-likelihoods is determined by the conditional entropy between the model input and target output. Based on these hypotheses, we presented an image-erasing strategy and UEN to assign and capture the conditional entropy distribution discrepancy between different ID datasets.
Our model, trained on a complex dataset, becomes transferable to other ID datasets.
Experimental results on the five datasets show that our method, without retraining, achieves comparable performance with the SOTA group-based OOD detection methods that require retraining on the ID datasets.


\clearpage
\section{Appendix}

\begin{sidewaystable}
\begin{tabular}{lcccclcccc}
\toprule
\multirow{2}{*}{ID/OOD} &\multicolumn{2}{c}{Ours($\mathcal{H}$)}& \multicolumn{2}{c}{Ours} & \multirow{2}{*}{ID/OOD} & \multicolumn{2}{c}{Ours($\mathcal{H}$))} & \multicolumn{2}{c}{Ours} \\
\cline{2-5}
\cline{7-10}
 & AUROC & AUPR & AUROC & AUPR & & AUROC & AUPR & AUROC & AUPR \\
\hline
CI10/CI100 &   $50.1\pm1.9$ & $49.4\pm0.3$ &   $\textbf{64.1}\pm\textbf{1.1}$ & $\textbf{52.3}\pm\textbf{0.3}$ & CI100/CI10 & $48.4\pm1.0$ & $49.8\pm0.2$ & $\textbf{62.2}\pm\textbf{1.5}$ & $\textbf{53.4}\pm\textbf{0.4}$ \\
\hline
An/Non-an & $\textbf{54.1}\pm\textbf{0.5}$ & $49.9\pm0.3$ & $53.3\pm2.2$ & $\textbf{49.9}\pm\textbf{0.1}$  & Non-an/An & $54.2\pm1.6$ & $51.6\pm0.3$ & $\textbf{55.3}\pm\textbf{1.6}$ & $\textbf{52.1}\pm\textbf{0.2}$ \\
\hline
C0/Ow(C0) & $77.3\pm1.1$ & $54.5\pm1.2$ & $\textbf{79.4}\pm\textbf{0.9}$ & $\textbf{55.3}\pm\textbf{0.4}$ & Ow(C0)/C0 & $63.5\pm1.5$ & $51.6\pm0.6$ & $\textbf{67.5}\pm\textbf{2.0}$ & $\textbf{52.4}\pm\textbf{0.5}$ \\
\hline
C1/Ow(C1) & $85.9\pm1.3$ & $\textbf{60.3}\pm\textbf{2.0}$ & $\textbf{86.2}\pm\textbf{1.1}$ & $59.3\pm0.3$ & Ow(C1)/C1 & $\textbf{87.8}\pm\textbf{0.7}$ & $\textbf{61.1}\pm\textbf{1.5}$ & $85.6\pm1.5$ & $59.8\pm0.3$ \\
\hline
C2/Ow(C2) & $\textbf{66.8}\pm\textbf{1.7}$ & $51.4\pm0.8$ & $62.9\pm0.1$ & $\textbf{52.4}\pm\textbf{0.1}$ & Ow(C2)/C2 & $\textbf{61.2}\pm\textbf{2.1}$ & $51.8\pm0.4$ & $59.7\pm2.4$ & $\textbf{53.8}\pm\textbf{0.3}$ \\
\hline
C3/Ow(C3) & $50.7\pm1.9$ & $49.4\pm0.4$ & $\textbf{51.1}\pm\textbf{2.0}$ & $\textbf{49.9}\pm\textbf{0.2}$ & Ow(C3)/C3 & $49.1\pm1.4$ & $50.1\pm0.3$ & $\textbf{49.4}\pm\textbf{1.2}$ & $\textbf{50.2}\pm\textbf{0.4}$ \\
\hline
C4/Ow(C4) & $\textbf{64.5}\pm\textbf{1.1}$ & $51.2\pm0.4$ & $64.2\pm1.9$ & $\textbf{53.5}\pm\textbf{0.6}$ & Ow(C4)/C4 & $62.4\pm1.7$ & $52.4\pm0.7$ & $\textbf{63.4}\pm\textbf{1.4}$ & $\textbf{54.6}\pm\textbf{0.2}$ \\
\hline
C5/Ow(C5) & $52.2\pm1.4$ & $\textbf{50.5}\pm\textbf{0.6}$ & $\textbf{52.6}\pm\textbf{1.9}$ & $50.3\pm0.2$ & Ow(C5)/C5 & $\textbf{56.1}\pm\textbf{0.7}$ & $\textbf{50.2}\pm\textbf{0.1}$ & $52.3\pm2.2$ & $50.1\pm0.3$ \\
\hline
C6/Ow(C6) & $50.4\pm1.6$ & $50.2\pm0.3$ & $\textbf{55.4}\pm\textbf{2.8}$ & $\textbf{50.3}\pm\textbf{0.4}$ & Ow(C6)/C6 & $\textbf{57.5}\pm\textbf{2.0}$ & $\textbf{51.2}\pm\textbf{0.5}$ & $50.9\pm2.4$ & $49.8\pm0.4$ \\
\hline
C7/Ow(C7) & $52.2\pm2.6$ & $50.3\pm0.5$ & $\textbf{57.0}\pm\textbf{1.4}$ & $\textbf{51.8}\pm\textbf{0.3}$ & Ow(C7)/C7 & $59.7\pm1.8$ & $\textbf{51.5}\pm\textbf{0.7}$ & $\textbf{64.3}\pm\textbf{1.2}$ & $51.2\pm0.3$\\
\hline
C8/Ow(C8) & $53.6\pm2.3$ & $50.5\pm0.4$ & $\textbf{64.5}\pm\textbf{2.4}$ & $\textbf{50.9}\pm\textbf{0.5}$ & Ow(C8)/C8 & $47.6\pm0.1$ & $49.6\pm0.3$ & $\textbf{67.0}\pm\textbf{1.2}$ & $\textbf{51.1}\pm\textbf{0.4}$ \\
\hline
C9/Ow(C9) & $62.9\pm0.1$ & $51.1\pm0.9$ & $\textbf{71.2}\pm\textbf{1.6}$ & $\textbf{60.0}\pm\textbf{0.5}$ & Ow(C9)/C9 & $70.4\pm0.6$ & $51.5\pm0.3$ & $\textbf{74.0}\pm\textbf{0.5}$ & $\textbf{60.1}\pm\textbf{0.2}$ \\
\bottomrule
\end{tabular}
\caption{ OOD detection results on CIFAR10 and CIFAR100 \cite{CIFAR10}. $\mathcal{H}$ refers to the real conditional entropy is fed into our Algorithm \ref{alg1}. CI10 and CI100 refer to the CIFAR10 and CIFAR100. An and Non-an refer to the  Animal and Non-animal categories in CIFAR10. C0 - C9 refer to the categories in CIFAR10 including airplane, automobile, bird, cat, deer, dog, frog, horse, ship and truck. Ow(Ci) refer to the other categories in CIFAR10 without Ci ($i \in \{0 - 9\}$).}
\label{Experiments_t8}
\end{sidewaystable}

\begin{figure*}[!h]
\centering
\includegraphics[width=17.5cm]{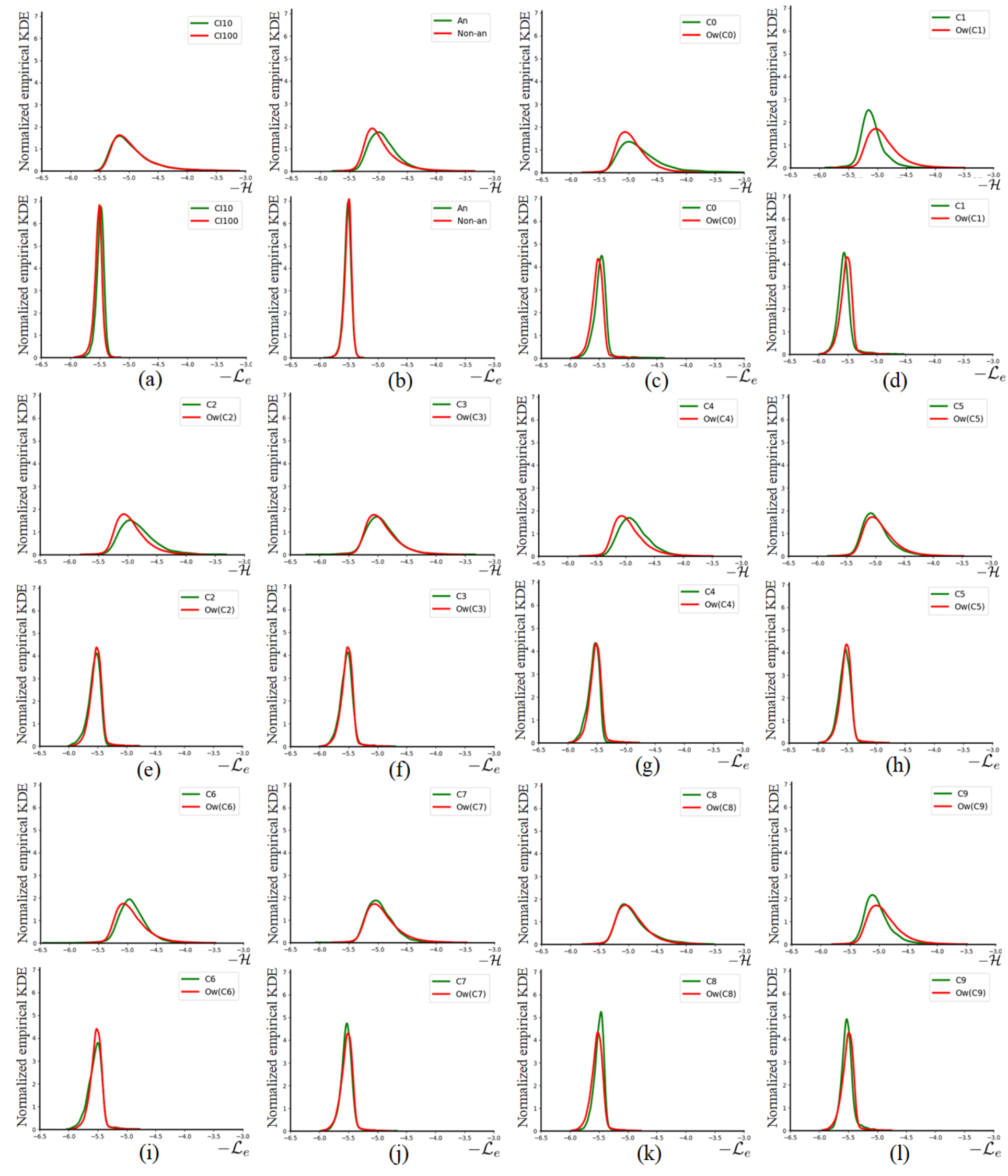}
\caption{The comparison between empirical distribution of model's negative log-likelihoods ($\mathcal{-L}_e$) and real negative conditional entropy ($\mathcal{-H}$) with different ID and OOD datasets.
The abbreviations of CI10, CI100, An, Non-an and C0 - C9 are the same as in Table \ref{Experiments_t8}.}
\label{Experiments_f6}
\end{figure*}

\subsection{A: Additional results on CIFAR10 and CIFAR100}
To further explore the potential of CETOOD, we conducted the following experiments based on CIFAR10 and CIFAR100 utilizing the \textit{Center} erasing strategy and the model pre-trained on ImageNet:
We conducted a total of three sets of experiments, with the experimental settings as follows:

1) \textit{CIFAR10 VS CIFAR100:}
Take the training set of CIFAR10 or CIFAR100 as ID dataset, and correspondingly, the testing set of CIFAR100 or CIFAR10 as OOD dataset for OOD detection.

2) \textit{Animal VS Non-animal:}
Images in CIFAR10 are divided into Animal and Non-animal categories. Then, take the training set of Animal or Non-animal as ID dataset, and correspondingly, the testing set of Non-animal or Animal as OOD dataset for OOD detection.

3) \textit{Ci VS Ow(Ci), $i \in \{1, \cdots, 10\}$:}
One category in CIFAR10 was randomly selected. Then, take the training set of the selected category or the rest categories as ID dataset, and correspondingly, the testing set of the rest categories or the selected category as OOD dataset for OOD detection.

The experimental results of the above three settings are tabulated in the Table \ref{Experiments_t8} (Ours).
We can observe from the Table \ref{Experiments_t8} that CETOOD cannot achieve effective OOD detection in all the three experiments.
To further analyze the reasons behind the poor performance of CETOOD, we calculate the real conditional entropy of ID data and OOD data for the three experiments and detect the OOD samples by feeding the real conditional entropy of samples into Algorithm \ref{alg1}. 
The OOD detection performance is shown in the Table \ref{Experiments_t8} (Ours($\mathcal{H}$)).
In addition, the empirical distribution of $\mathcal{-H}$ and $\mathcal{-L}_e$ of the ID data and OOD data in the three experiments are shown in Figure~\ref{Experiments_f6} for an intuitive comparison.

\begin{figure*}[!t]
\centering
\includegraphics[width=17cm]{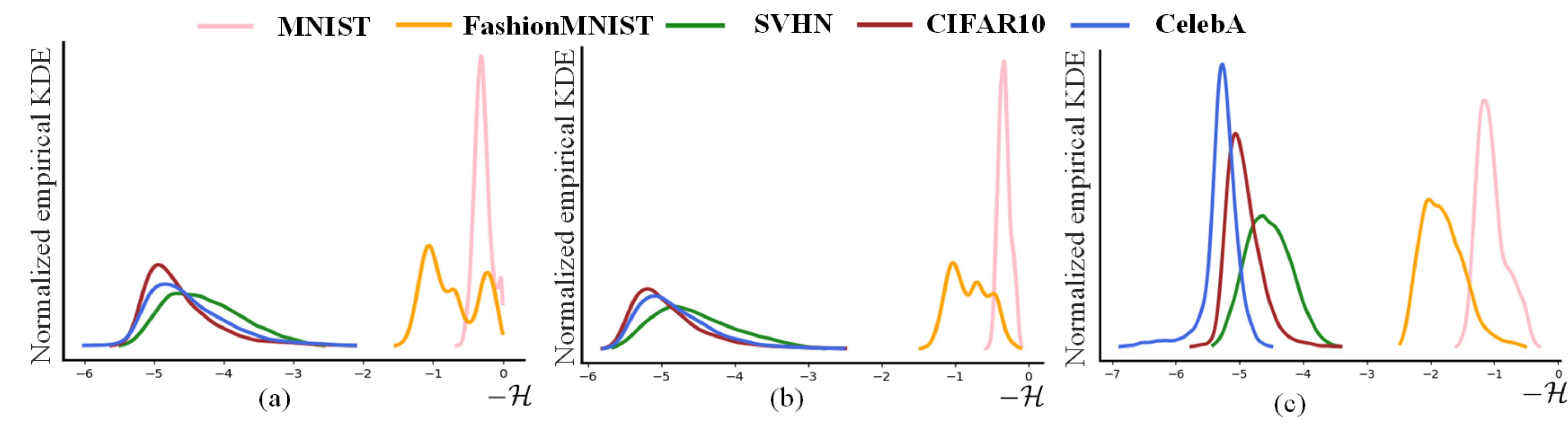}
\caption{For different datasets and image-erasing strategies, the empirical distribution of negative conditional entropy ($\mathcal{-H}$) of the erased image patch with the surrounding information as condition.
(a) The average results of four scenarios in strategy Corner.
(b) The average results of four scenarios in strategy Side.
(c) The results of strategy Center.}
\label{Experiments_f3}
\end{figure*}

Firstly, we can observe from Figure~\ref{Experiments_f6} that CETOOD cannot detect OOD data effectively because the image-erasing strategy (\textit{Center}) cannot assign exclusive conditional entropy distribution for the ID and OOD datasets, in 3 experimental settings.
Therefore, even though our model can effectively capture conditional entropy, it still performs poorly in 3 experimental settings.
This shortcoming can be alleviated by redesigning the image-erasing strategy.
An adaptive image-erasing strategy can be further investigated to address the limitation.
For example, we can maximize the conditional entropy between the surrounding information and the erased patch by erasing the regions of semantic information concentration based on the semantic segmentation task, which will be explored in the future.
Secondly, as shown in Figure~\ref{Experiments_f6}, the variance of the log-likelihoods is significantly smaller than that of the real conditional entropy distribution. 
Meanwhile, the results in Table \ref{Experiments_t8} show that the OOD detection performance of our model is better than detecting OOD samples using real conditional entropy.
The above experimental results demonstrate that $\mathcal{L}_e$ in the loss function can effectively red uce the variance of the image generation posteriori distribution of the similar sample, thus improving the discriminability of the captured image generation posteriori distribution compared with the original conditional entropy
distribution.

\subsection{B: Visualization of Conditional entropy distribution of different image-erasing strategies}

To investigate how the model is trained with the strategy \textit{Center} achieves unbalanced improvements, we calculate the real conditional entropy of all datasets with different image-erasing strategies.
The empirical distribution of negative conditional entropy ($\mathcal{-H}$) is shown in Figure~\ref{Experiments_f3}. 
For every test sample, we calculate the average negative conditional entropy of all scenarios that belong to the same image-erasing strategy as its final representation.
From Figure~\ref{Experiments_f3}, it can be seen that the strategy Corner and Side cannot assign exclusive conditional entropy distribution for CIFAR10 and CelebA, and the conditional entropy distribution between them is almost inseparable.

\subsection{C: Visualization of Log-likelihoods distribution of different training set}

\begin{figure}[!h]
\centering
\includegraphics[width=9cm]{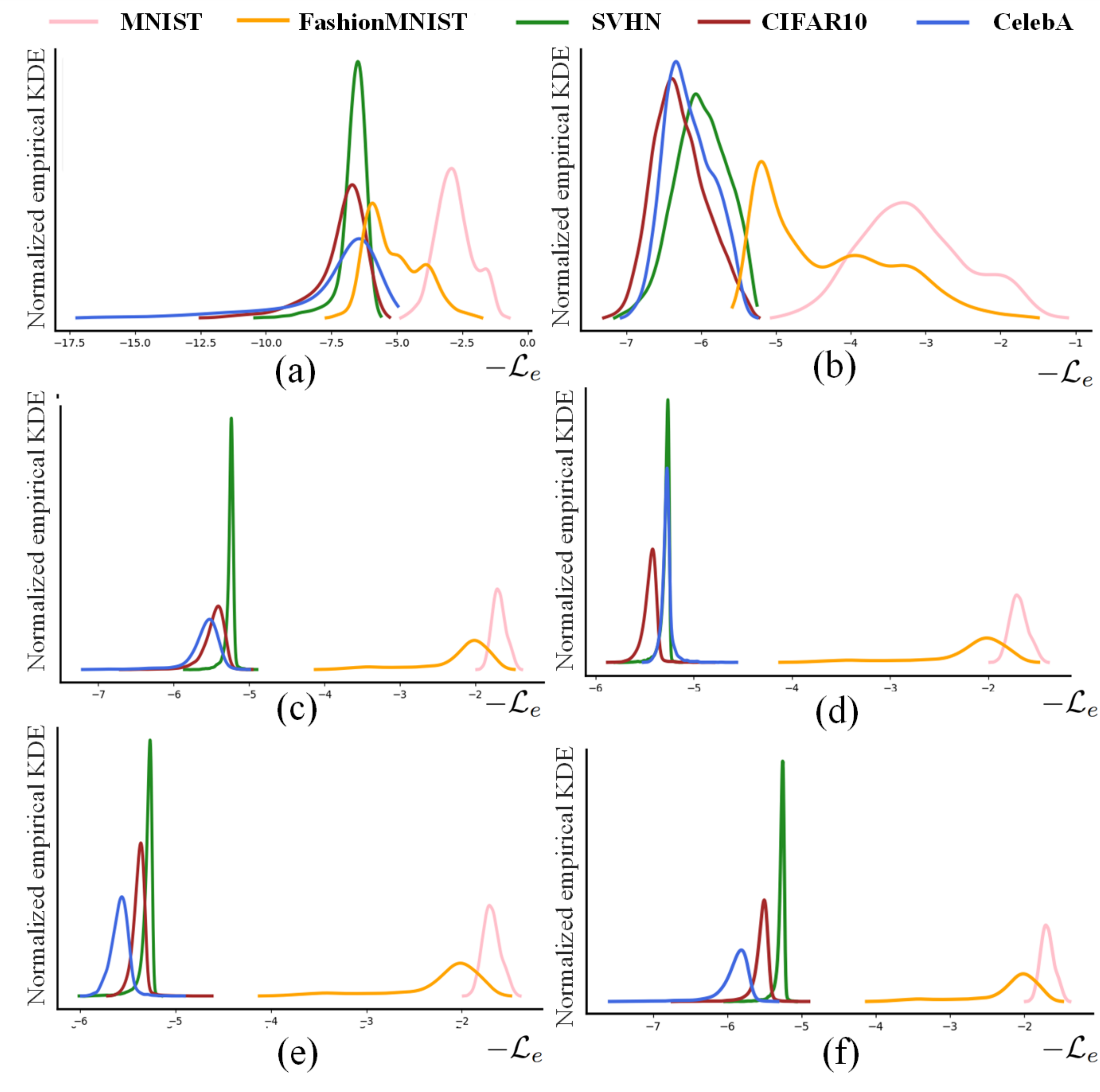}
\caption{The empirical distribution of log-likelihoods of the erased patch when the model is trained with different training set, including: MNIST(a), FashionMNIST (b), SVHN (c), CelebA (d) and CIFAR10 (e) and ImageNet(f).}
\label{Experiments_f5}
\end{figure}

To analyze the effect of the Training Set, we use the training sets of 6 datasets, including MNIST, F-MNIST, SVHN, CelebA, CIFAR10 and ImageNet, to train our model respectively.
In order to intuitively demonstrate how the complexity of training data affects the model's ability of conditional entropy capturing, in Figure~\ref{Experiments_f5}, we give the model's log-likelihoods when it is trained with different ID datasets.

We can observe from Figure~\ref{Experiments_f5} that the OOD detection performance of the model gradually improves as the complexity of training data increases.
As the several models in the Figure~\ref{Experiments_f5} are trained with the same image-erasing strategy, the conditional entropy distributions of the 5 ID datasets are consistent among 6 experiment settings, and exclusive conditional entropy distribution is assigned to different ID datasets.
Therefore, the experimental results show that the representation power of a model to capture the conditional entropy is affected by the complexity of training data. 
Highly complex training data helps the model better capture the conditional entropy of generating the erased patch from its surrounding in the ID sample.
In addition, according to the Motivation section, conditional entropy determines the lower bound of the negative-log-likelihoods. Our model can capture the conditional entropy effectively, which indicates that the posteriori of image generation in this model is close to MAP.
Therefore, the experimental results in Figure~\ref{Experiments_f5} are consistent with the conclusion of observation (ii), that is, the DGMs learn low-level features rather than semantic information, so a pre-trained DGM-based OOD detection model can transfer among different ID datasets.

\end{document}